\documentclass{article}

\usepackage[utf8]{inputenc}

\usepackage[numbers]{natbib}
\usepackage{graphicx}

\usepackage{caption}
\usepackage{subcaption}

\usepackage{comment}

\usepackage{authblk}
\usepackage{booktabs,subcaption,amsfonts,dcolumn}

\usepackage{url}
\usepackage [autostyle, english = american]{csquotes}
\MakeOuterQuote{"}

\title{Odor Descriptor Understanding through Prompting}
\author{Laura Sisson  \\ laurasisson@google.com}
\affil{Google}
\usepackage{makecell}
\clearpage
\usepackage{float}
\usepackage{hyperref}

\newcommand{\desc}[1]{{\small``#1''}}
\newcommand{\prompt}[1]{{\small\textit{``#1''}}}

\begin{document}

\maketitle

\section{Abstract}
Embeddings from contemporary natural language processing (NLP) models are commonly used as numerical representations for words or sentences. However, odor descriptor words, like \desc{leather} or \desc{fruity}, vary significantly between their commonplace usage and their olfactory usage, as a result traditional methods for generating these embeddings do not suffice. In this paper, we present two methods to generate embeddings for odor words that are more closely aligned with their olfactory meanings when compared to off-the-shelf embeddings. These generated embeddings outperform the previous state-of-the-art and contemporary fine-tuning/prompting methods on a pre-existing zero-shot odor-specific NLP benchmark.

\section{Introduction}
Word embeddings from NLP models contain semantic knowledge that can be used across disparate tasks. However, in certain contexts, simply using the hidden layer activations for a particular word or phrase may not extract the desired knowledge from the NLP model. This is the case for odor descriptors, where their meanings in olfaction differ from everyday-use meanings. 

Most NLP models are trained on general corpora and so their internal representations for words represent the most generalized context a word appears in. For example, in everyday speech \desc{leather} refers to the material from which certain clothes are made of, or perhaps the tough, textured \desc{leathery} feel of that material. This means that the off-the-shelf NLP model embedding for \desc{leather} will likely be similar to words like \desc{jacket}, \desc{rugged} or \desc{hide}. Though the term \desc{leather} in olfaction refers to the scent of that material, it is perceptually similar to seemingly unrelated words like \desc{musk} or \desc{amber}\cite{noauthor_undated-mg}.

\begin{figure}[h]
	\centering
	\includegraphics[scale=.5]{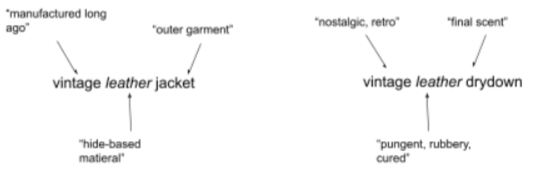}
	\caption{\small\textbf{Connotations for individual words vary based on context.}}
	\label{fig:connotations}
\end{figure}

The connotations in odor are abstract and aesthetic, and so are hard to pin down numerically. Overall though, the embeddings for odor descriptors in similar categories (take for example \desc{apple}, \desc{fruit}, and \desc{peachy}) should be closer to each other than the embeddings for words in dissimilar categories (like \desc{gasoline}).

In practice, this means that the same procedures that previously resulted in useful embeddings for other domains may not function when applied to odor descriptors. To aggravate the issue, though odor may be a common topic of discussion, datasets specifically focused on odor\cite{Keller2016-ci} are small in comparison to the hundreds of terabytes of data contained in contemporary NLP datasets\cite{noauthor_undated-ir}.

\subsection{Previous Work}

A variety of work on quantitative analysis has been conducted on odor descriptors\cite{Zarzo2021-rb}. 
One of the most innovative approaches to NLP embeddings in this space is “Predicting natural language descriptions of mono-molecular odorants”\cite{Gutierrez2018-hh}. The researchers used the fasttext\cite{Bojanowski2016-og} model trained on Wikinews\footnote{Available at https://fasttext.cc/docs/en/english-vectors.html} to generate 300 dimensional embeddings for a set of 150 single-word descriptors across 58 molecules from the \textit{DREAM}\cite{Keller2017-og} and \textit{Dravnieks}\cite{Dravnieks_A_ASTM_Committee_E-18_on_Sensory_Evaluation_of_Materials_and_Products1985-cx} odorant datasets.

In order to evaluate the generated word embeddings, the researchers proposed a zero-shot task where ratings on the 19 descriptors used in the \textit{DREAM} dataset are used to predict the ratings on the 131 descriptor used in the \textit{Dravnieks} dataset for the same molecule. Both sets of ratings are in the [0,100] range. A linear regression model from scikit-learn\cite{Pedregosa2011-mu} is fit between the 19x300 and 131x300 embeddings, and then predicts the 131 \textit{Dravnieks} ratings from the 19 \textit{DREAM} ratings.

One caveat not mentioned in the original study is the difficulty in generating embeddings for multi-word or out-of-dictionary descriptors using fasttext. In order to get around this issue, the researchers manually cleaned the descriptor lists. Though the \textit{DREAM} dataset contains ratings on both \desc{urinous} and \desc{cat urine}, both of these descriptors are simplified to \desc{urine}, which is within the dictionary for fasttext. Some descriptors, like \desc{cleaning fluid} were removed altogether from the \textit{DREAM} dataset.

As a result, though the original task described is a 19 to 131 rating prediction task, the full \textit{DREAM} dataset is 146 descriptors. For the rest of this paper, the complete 19x146 descriptor task will be referred to as the full-descriptor task, and the previously described 19x131 task will be referred to as the single-word task.

\subsection{Our Work}
The contributions of our paper are as follows: Firstly, we highlight the need for odor-specific techniques of embedding generation, which contextualize the embedding to the olfactory space, in comparison with off-the-shelf methods which instead reflect the everyday-use meaning. Secondly, we provide two such techniques that better capture the olfactory connotations of these words, outperforming previous work and a variety of contemporary methods. For future researchers, our state-of-the-art method and the resultant embedding prompts are open-source\footnote{Available at https://github.com/laurahsisson/odor-prompting}.

\section{Methods}

\subsection{Randomized Baseline}
In order to provide a baseline to these tasks, we generated random embeddings of dimension = 300 for each of the 150 descriptors. Fitting the linear regression model across these random vectors resulted in a score of .25 on the single-word task and a score of .26 on the full-descriptor task. Increasing the dimensionionality of these vectors to did not result in an increase in score.

\subsection{Advances in NLP}
Many larger and more powerful models have been developed since fasttext’s release. Though cutting-edge models like LaMDA\cite{Thoppilan2022-fu} and GPT-3\cite{Brown2020-og} have resulted in impressive scores across a variety of natural language benchmarks, these models may be prohibitively expensive to train for the average researcher. Though BERT\cite{Devlin2018-ke}, a moderate-sized transformer\cite{Vaswani2017-wh} model, is nearly 3 years old, it strikes a balance between performance and resource-usage. In this case the publicly available BERT-large-uncased from Hugging Face’s Transformers\cite{Wolf2019-te} is powerful enough while still feasibly trained within our budget.

\subsection{Simple Approaches with BERT}
A trivial approach to generate word embeddings is to use BERT's wordpiece dictionary. To extract the wordpiece embedding for \desc{flowery}, the descriptor is fed into BERT's tokenizer and split into \desc{flow\#\#} and \desc{\#\#ery}. Averaging across the vector representations of these two wordpieces in BERT’s pretrained dictionary provides an embedding for \desc{flowery}.

A more in-depth approach is to leverage the hidden layer embeddings for the isolated descriptors. Though BERT is a contextual model, it is possible to generate word embeddings for the odor descriptors without a context. To do this, the descriptor of interest (\desc{flowery}) is fed into the model, and the averaged hidden layer activations are used as 1024 dimensional embeddings. For the rest of the paper, the layer from which to extract activations is treated as a hyperparameter. 

\subsection{Context Development}
There have been numerous techniques developed for generating contexts, now called prompts, in modern NLP research, ranging from human generated guesses\cite{Schick2020-ks} to fine-tuning approximate contexts\cite{Liu2021-qu}. Usually, these prompts take the form of words or tokens surrounding a \textit{[blank]} where the descriptor of interest is substituted, but sometimes the prompts are represented purely as differentiable vectors\cite{Lester2021-vg}, which are fed directly into the model’s hidden layers.

We leveraged a variety of different approaches, not all of which were successful. For elaboration on these unsuccessful techniques, see the \hyperref[sec:failed]{“Failed Approaches”} section in the appendix. We present two successful approaches: decent results with human generated prompts, and a new state-of-the-art on the aforementioned benchmark using a prompt mining technique. 

\subsection{Human Generation}
Human generated prompts involve brainstorming domain-specific contexts for the descriptor of interest. Though this method is not rigorous or rooted in statistical theory, it is simple to implement and computationally inexpensive. For the task of embeddings in the odor space, example human generated prompts are \prompt{[blank] scent} or \prompt{smells like [blank]}. Another benefit of this method is that no corpus needs to be collected and there is no additional training or fine-tuning of the model, so the computational cost is low.

\subsection{Corpus for Mining}
Prompt mining\cite{Jiang2019-fd} involves searching domain relevant corpora for the contexts surrounding the descriptors of interest. For the corpora from which to mine contexts, we used the \textit{GoodScents}\cite{noauthor_undated-wj}
perfume materials database and \textit{Leffingwell PMP 2001}\cite{noauthor_undated-zv} database. These catalogs consist of lists of chemicals with short olfactory descriptions, generated by experts. 

Prior to any cleaning, these two corpora totalled 650,000 words when combined. The dataset required standardization, as some chemical descriptions consisted of lists of descriptors (\desc{musky}, \desc{sweet}, \desc{chalky}) and others were closer to full sentences (\desc{Ambrofix is a highly powerful, highly substantive and highly stable ambery note for use in all applications}). We aimed to isolate and list the odor descriptors: in other words, \desc{musky}, \desc{sweet} and \desc{chalky} would all be separated, whereas \desc{highly powerful, highly substantive and highly stable ambery note} would be kept together as a multi-word descriptor. 

In order to canonicalize the input, we transformed the dataset into a comma separated list of word chunks, parsed using spaCy\cite{Honnibal2017-pv}. This resulted in around 40k unique total descriptors. The median and mode word occurrence frequency was 1, but the mean word occurence frequency was around 10. 

The following descriptors from the single-word task did not appear at all in the dataset: 
\\ 
\desc{peppers}, \desc{mothballs}, \desc{weeds}, \desc{grainy}, \desc{beery}, \desc{rope}, \desc{stale}, \desc{potatoes}, \desc{peanuts}, \desc{carbolic}, \desc{kipper}, \desc{semen}, \desc{sooty}, \desc{methane}, \desc{sewer}, \desc{cadaverous}, \desc{sickening}, \desc{chalky}, \desc{cold}, \desc{musky}, \desc{decayed}

\begin{figure}[h]
	\centering
	\includegraphics[scale=.5]{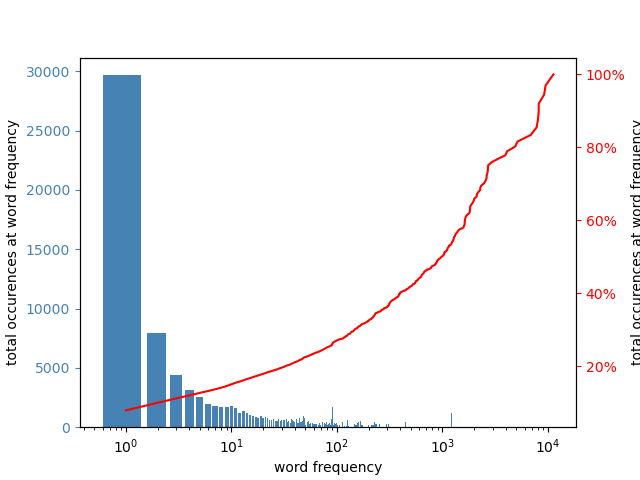}
	\caption{\small\textbf{Density of word distribution at different frequencies.} Each bar represents the sum total of all occurrences for words that appear that many times. In other words, the first bar represents the total count of all words in the dataset that appear only once.}
	\label{fig:frequency}
\end{figure}

The Pareto Chart (Figure \ref{fig:frequency}) represents the distribution of all descriptors in the dataset. 

Though a large number of descriptors appear only once, when we removed all these singleton descriptors, only 7\% of the total dataset was discarded. The curve of the cumulative sum suggests that there is a relatively small set of descriptors that the authors of these datasets use frequently, and they reach for niche descriptors only occasionally.

\begin{table}[h]
\small
\begin{tabular}{ |p{2.5cm}|p{2.5cm}|p{2.5cm}|p{2.5cm}| p{2.5cm}|  }
	\hline
	\makecell[lt]{Highest \\ Frequency} & \desc{musk}, 11354       & \desc{bergamot}, 9542                   &  \desc{vanilla}, 8313       \\
	\hline
	\makecell[lt]{Middle \\ Frequency}  & \desc{cotton blossom}, 7 & \desc{orangeblossom}, 7            & \desc{melilot}, 6            \\
	\hline
	\makecell[lt]{Lowest \\ Frequency}  & \desc{captive use}, 1    & \desc{fresh fruity green melon}, 1 & \desc{natural wax}, 1 \\
	\hline
\end{tabular}
\caption{\label{tab:frequencies} \textbf{Sampling of descriptors} in different frequency bands.}
\end{table}

For further cleanup, we then merged descriptors if their edit distances and embeddings were close enough (\desc{chocolatey} and \desc{chocolate}). We chose the canonical form of the pair based on frequency. From there, we had a dataset of around 2,000 unique descriptors, ready for mining.

\begin{figure}[h]
	\centering
	\includegraphics[scale=.5]{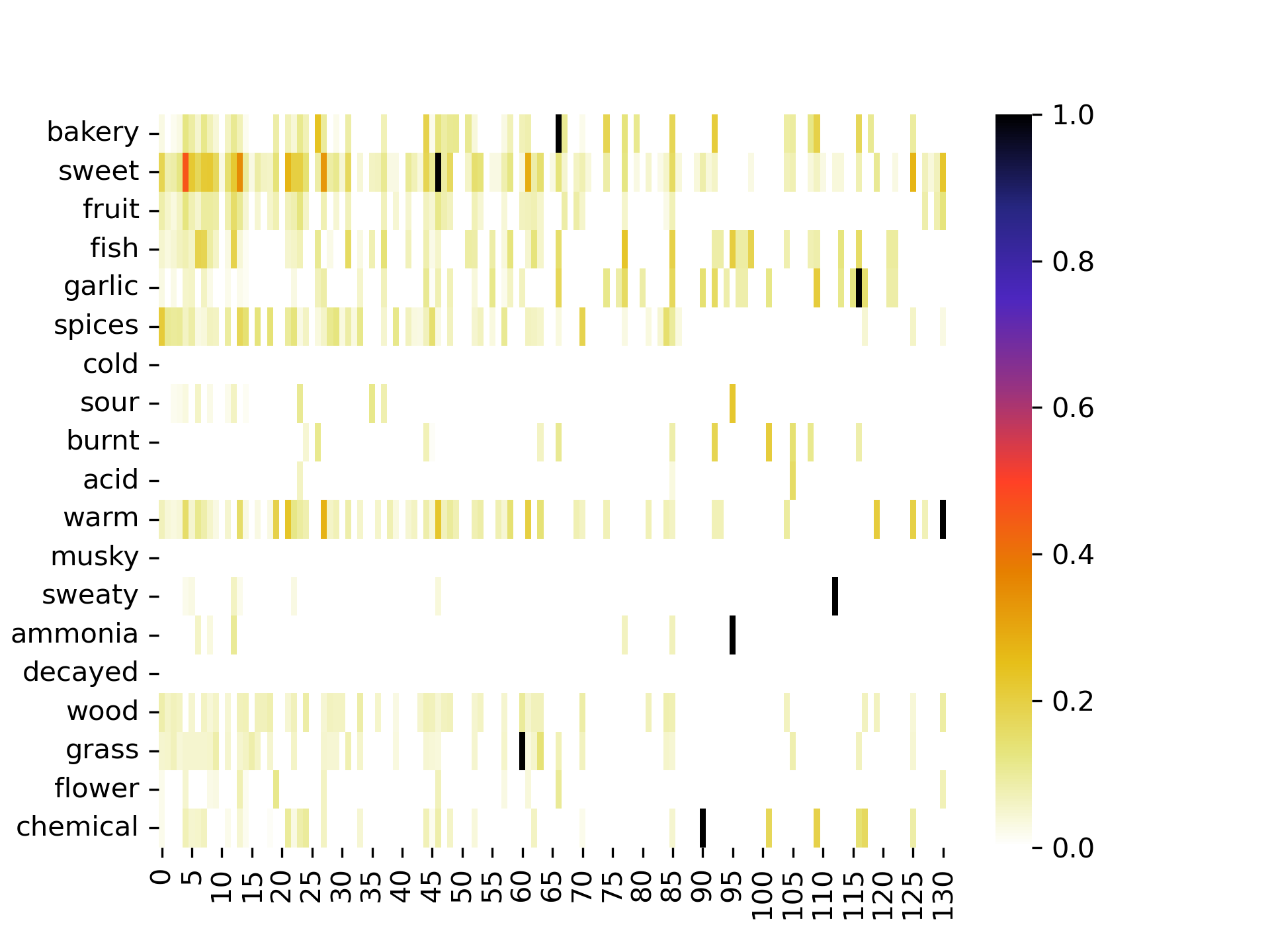}
	\caption{\small\textbf{Co-occurrence frequency in our dataset between the 19 descriptors from the \textit{DREAM} dataset (labeled y axis) and the 131 descriptors from the \textit{Dravnieks} dataset in the single-word task (enumerated x axis).} Note that \desc{bakery}, \desc{sweet}, \desc{garlic}, \desc{warm}, \desc{sweaty}, \desc{ammonia}, \desc{grass} and \desc{chemical}
 appeared in both datasets, so there are perfect 1.0 correlations for the intersections of these descriptors with themselves.}
	\label{fig:cooccurence}
\end{figure}

\subsection{Mining Algorithm}
To mine the prompt, we used a k-beam search on the generated corpus. The algorithm starts initially with an empty prompt (\prompt{[blank]}), and then for each beam, a random descriptor from the corpus is selected, weighted based on frequency. This descriptor is then added arbitrarily as either a new suffix or prefix to the prompt. Each generation contains \textit{k} beams (75 in our trials), and at subsequent iterations, only the top \textit{k} most promising beams are kept. The beams are evaluated based on their scores averaged between the single-word and full-descriptor tasks. This beam search works recursively, so that in the second and subsequent generations many potential prompts are explored (75*75=5625 in our trials).

\begin{figure}[h]
	\centering
	\includegraphics[scale=.5]{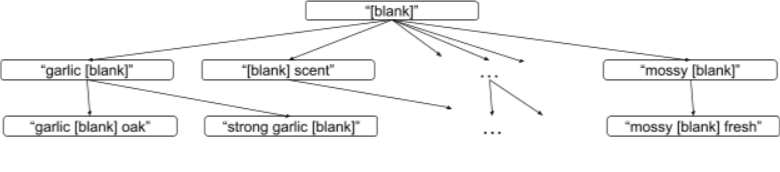}
	\caption{\small\textbf{Iterative construction of mined prompt.} Each depth represents a generation, from which only a select number of prompts are used as bases for the next generation.}
	\label{fig:kbeam}
\end{figure}

\section{Results}
When using the hidden layer activations for the unprompted descriptor words alone, BERT-large-uncased achieves a score of .44 on the single-word task, slightly lower than the older fasttext model, and a score of .39 on the full-descriptor task (hidden layer = 6).

\begin{figure}[h]
	\centering
	\includegraphics[scale=.5]{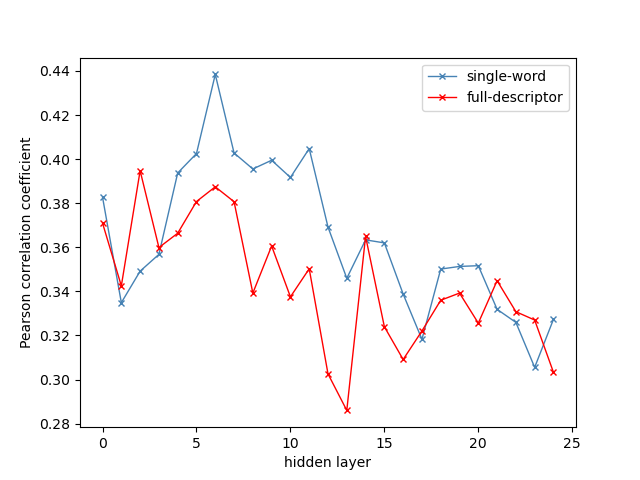}
	\caption{\small\textbf{Performance of BERT without a prompt on the single-word and full-descriptor tasks}, as a function of which hidden layer is used as embeddings.}
	\label{fig:layers}
\end{figure}

The most successful human generated prompt was \prompt{essence [blank] flavored} (hidden layer = 12), which scored .46 on the single-word task and .49 on the full-descriptor task, improving on the off-the-shelf BERT model without adding significant computational resources.

For our mined prompt, the k-beam search ran for 25 generations, though the score did not improve after the 14th generation, producing the best prompt of \prompt{many woody fresh butter popcorn woody orange biscuit rubbery [blank] intense garlic woody comments warm evaluation comments umami grape rain one”} (hidden layer = 11) with a score of .55 on the single-word task and a score of .56 on the full-descriptor task.

\begin{table}[h]
\small
\begin{tabular}{ |p{4cm}|p{3.5cm}|p{2.5cm}|}
	\hline
	Model                     & single-word & full-descriptor \\
	\hline
	random baseline           & .25         & .26             \\
	\hline
	fasttext  & 0.47        & 0.00            \\
	\hline
	\makecell[lt]{BERT \\ (wordpiece dictionary)} & 0.36        & 0.34            \\
	\hline
	\makecell[lt]{BERT \\ (unprompted)}         & 0.44        & 0.39            \\
	\hline
	\makecell[lt]{BERT \\ (human prompt)}       & 0.46        & 0.49            \\
	\hline
	\makecell[lt]{BERT \\ (mined prompt)}       & 0.55        & 0.56            \\
	\hline
\end{tabular}
\caption{\label{tab:comparison} \textbf{Comparison of performance across models.}}
\end{table}

For the breakdown of improvement by descriptor, see supplementary information. For a summary, Table \ref{tab:losses} contains the descriptors with the largest gains and losses.

\begin{table}[h]
\small
\begin{tabular}{ |p{3cm}|p{3cm}|p{3cm}|}
	\hline
	Descriptor      & Change in score & Occurrences \\
	\hline
	\desc{mouse}     & .92         & 1                          \\
	\hline
	\desc{perfumery} & .87         & 60                         \\
	\hline
	\desc{blood}     & .82         & 11                         \\
	\hline
	\hline
	\desc{raisins}    & -0.89       & 4                          \\
	\hline
	\desc{grapefruit} & -0.75       & 2                          \\
	\hline
	\desc{cantaloupe} & -0.74       & 23                         \\
	\hline
	
\end{tabular}

\caption{\label{tab:losses} \textbf{Descriptors with largest changes in accuracy between fasttext and BERT with our mined prompt}, tabulated with their total occurences in our cleaned dataset.}
\end{table}

\begin{figure}[h]
	\caption{PCA embedding space across different methods}
	\begin{subfigure}{0.32\textwidth}
		\includegraphics[width=\linewidth]{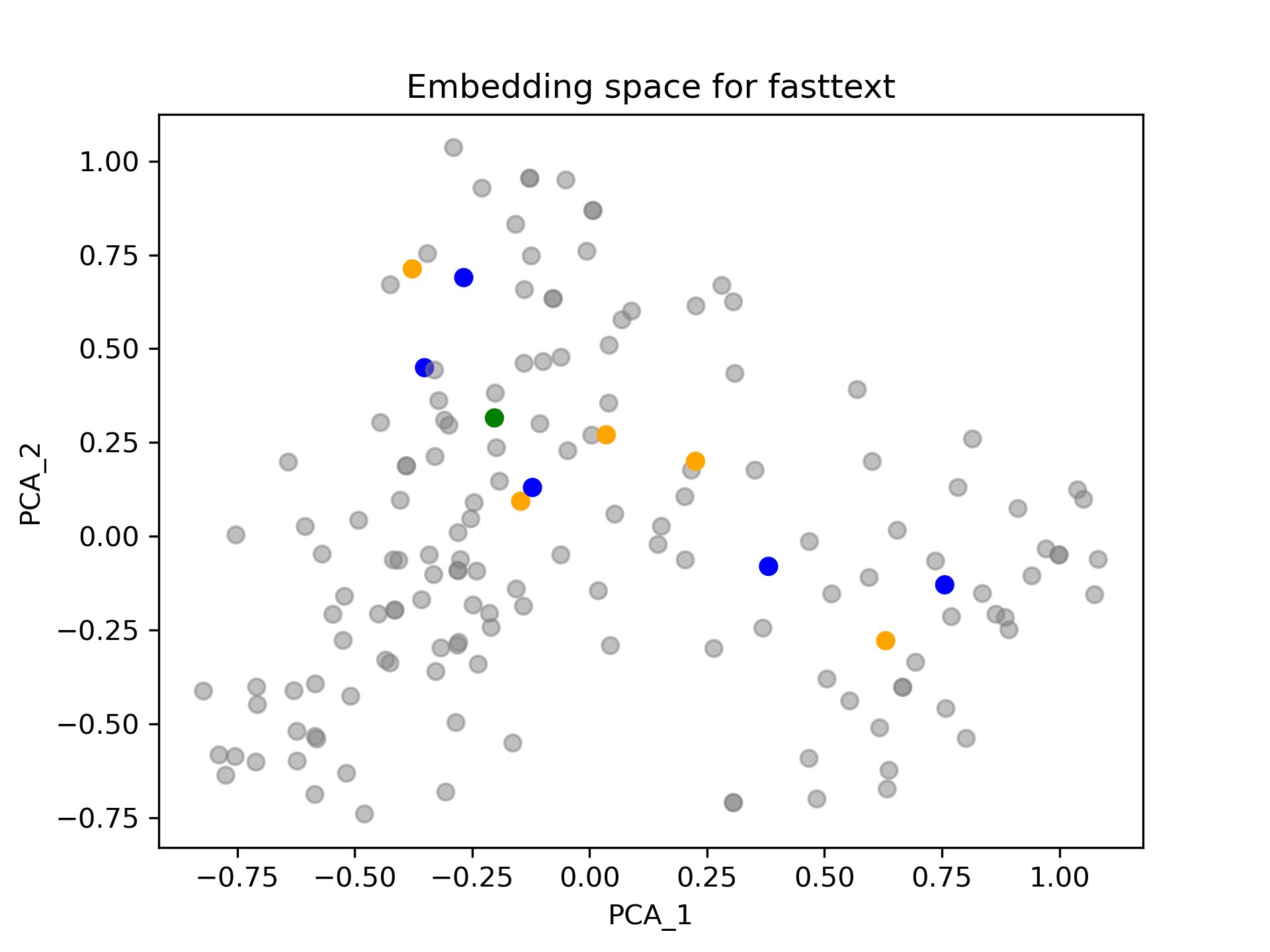}
		\subcaption{\small\textbf{Embedding space for fasttext.} Embedding space is confined, so the negatives are as close or closer to “leather” than the positives.}
		\label{fig:fasttext}
	\end{subfigure} \hspace{.75mm}
	\begin{subfigure}{0.32\textwidth}
		\includegraphics[width=\linewidth]{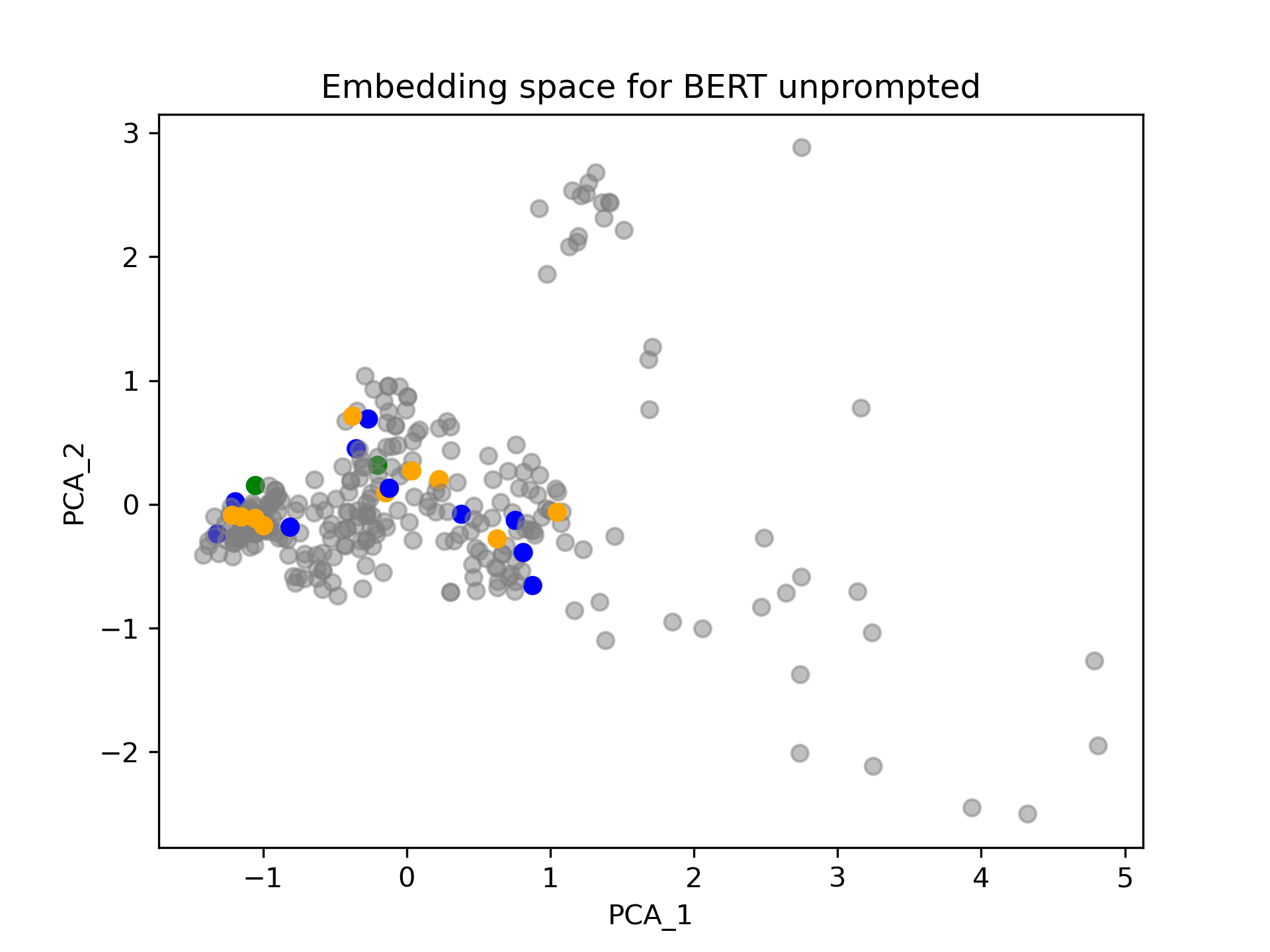}
		\subcaption{\small\textbf{Embedding space for BERT unprompted.} Embedding space is spread out, and the central cluster is disorganized, with most negatives close to “leather”.
}
		\label{fig:unprompted}
	\end{subfigure} \hspace{.75mm}
	\begin{subfigure}{0.32\textwidth}
		\includegraphics[width=\linewidth]{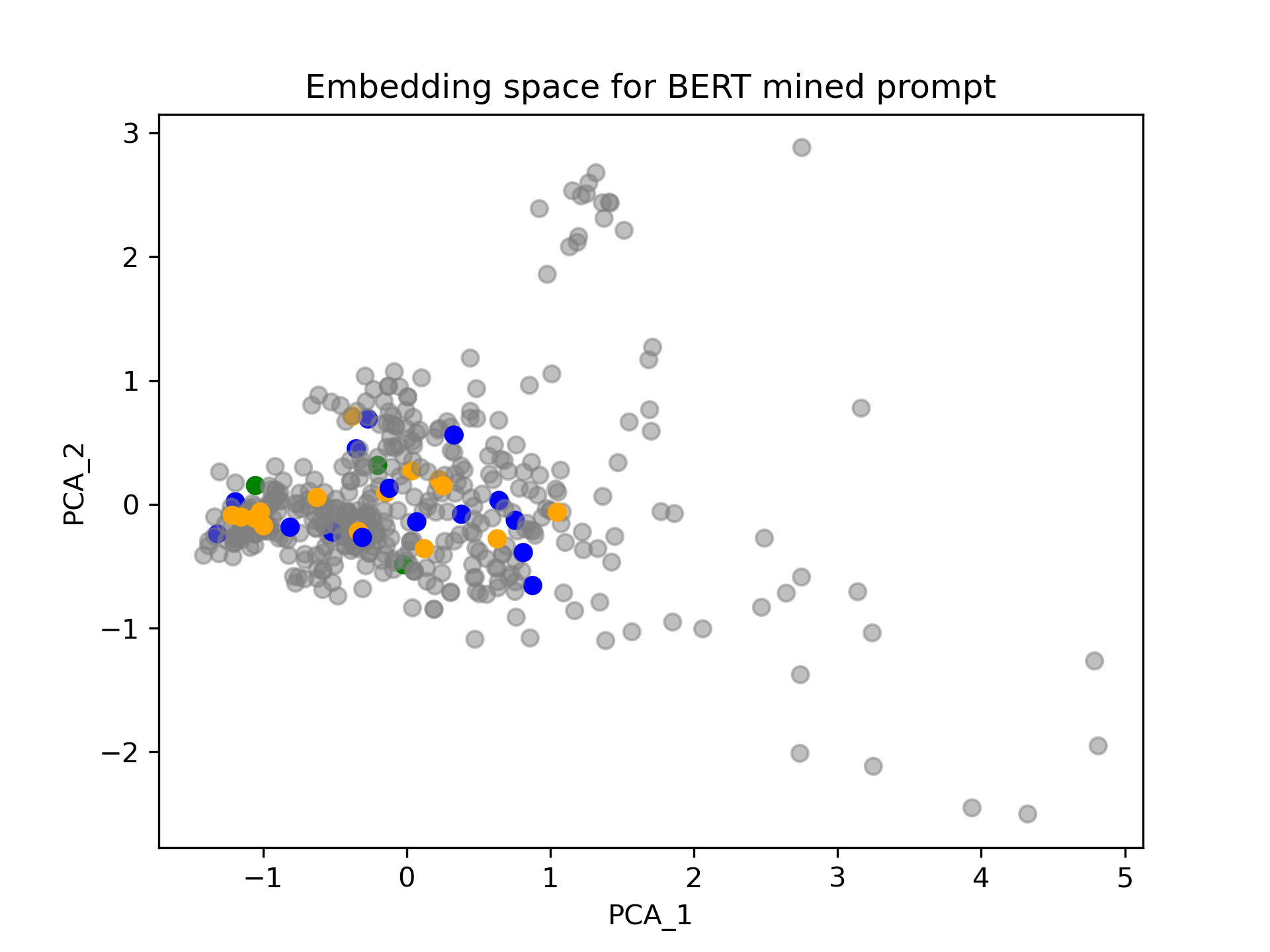}
		\subcaption{\small\textbf{Embedding space for BERT with the mined prompt.} Embedding space is relatively, neither too dense nor too sparse. }
		\label{fig:minedprompt}
	\end{subfigure}
	\label{fig:embeddings}
\end{figure}

For a visual comparison, we present the embedding spaces of various techniques (Figure \ref{fig:embeddings}) reduced to 2 dimensions through principal component analysis (PCA). To highlight the shift in the embedding space, \desc{leather} is charted in green. Five odor descriptors related to \desc{leather} are charted in blue (\desc{musky}, \desc{gasoline}, \desc{smoky}, \desc{amber}, \desc{musk}). Five additional words which are similar in the common usage to \desc{leather} but not related to odor are included as negatives and are charted in orange (\desc{jacket}, \desc{rugged}, \desc{hide}, \desc{material}, \desc{tanning}").

BERT with the mined prompt is the only embedding space where the negatives are on average further from “leather” than the positives, using Euclidean distance. To measure the spread of the embedding space, the average distance from each embedded point to the centroid of all points is also provided in Table \ref{tab:pca}. As can be seen in the PCA figures, the embedding space for BERT without a prompt is more spread out than the other two methods.

\begin{table}[h]
\small
\begin{tabular}{ |p{2.5cm}|p{2.5cm}|p{2.5cm}|p{2.5cm}|}
	\hline
	Model               &   \makecell[lt]{Mean distance \\ to Centroid} &   \makecell[lt]{Negatives \\to Leather} &  \makecell[lt]{Positives \\to Leather} \\
	\hline
	fasttext              & 1.94                 & 2.40                 & 2.64                 \\
	\hline
	\makecell[l]{BERT \\ (unprompted) }  & 3.68                 & 3.82                 & 4.29                 \\
	\hline
	\makecell[l]{BERT \\ (mined prompt)} & 1.20                 & 1.40                 & 1.31                 \\
	\hline
\end{tabular}
\caption{\label{tab:pca} \textbf{Quantitative comparison of embedding spaces} of fasttext and our two methods. }
\end{table}

\section{Conclusion}
Mining a prompt through a k-beam search on a domain specific dataset outperforms the randomized baseline and the previous state-of-the-art. Though the more recent BERT model struggled off-the-shelf, careful selection of a prompting algorithm meant large gains on the single-word task and a strong score on the full-descriptor task. It is also interesting to note that even using a simple human generated prompt produces decent results. 

The best prompt we found ("many woody fresh butter popcorn woody orange biscuit rubbery [blank] intense garlic woody comments warm evaluation comments umami grape rain one”) is nonsensical and contains duplicate words, as well as words not obviously related to olfaction, yet it outperforms human generated prompts containing tokens more superficially related to odor. Further research into why seemingly irrelevant words function well as prompt tokens is necessary, but it may suffice to imagine that what humans consider relevant differs from what has been encoded into textual usage as relevant: the deeper meaning eludes a surface level analysis. 

We offer a tested context for researchers in the odor space looking to generate numerical representations for odor words. For example, researchers often need to unify descriptions in list form (\desc{musky, sweet, chalky}) with descriptions in plaintext (\desc{Ambrofix is a highly powerful…}) or in rating form (\desc{musky: 50, flowery: 39}) while preserving each description’s meaning: future researchers can use our embeddings to convert all these different descriptions into vectors. Another application we are particularly interested in seeing is determining a core set of odor descriptors on which olfaction experts rate new odorants, perhaps generating using a clustering approach on the descriptor embeddings.

\subsection{Additional Considerations}
This paper was affected by the limited size of odor specific NLP corpora. As stated before, though other domains have datasets consisting of billions of words\cite{noauthor_undated-ir}, we struggled to collect half a million words. Even then, this dataset came out of private and confidential catalogs, meaning distribution to other researchers is infeasible.

However, there are actually large repositories of odor descriptions online, in the form of perfume discussion forums like Fragrantica\cite{noauthor_undated-dp} and Basenotes\cite{noauthor_undated-ub}. Unfortunately, these are legally guarded, and contain programmatic blocks against web crawlers. In practice, this content may be less specific than Goodscents or Leffingwell, as they contain freeform writing by untrained contributors, and the olfactory descriptions are mixed in among non-odor discussion about sales, packaging and branding. However, these datasets are orders of magnitude larger than the chemical repository datasets, and with some cleaning, these comments and reviews would be highly useful to future researchers. 

Finally, the evaluation benchmark was limited in scope. It involved only 150 descriptors, out of thousands, rated on a small number of examples. For a more realistic odor-specific NLP task, something akin to the Microsoft Research Paraphrase Corpus\cite{Lan2017-nq} for olfaction could be created. For evaluation, a model would be tasked to determine whether or not two olfactory descriptions were describing the same odorant or different odorants. The data required to generate thousands of such examples for this task is already available on the internet, and only needs to be cleaned up and made publicly available. Without this, NLP for olfaction will unnecessarily remain a low data domain.

\subsection{Acknowledgements}
We thank Dr. Alex Wiltschko and Dr. Benjamin Sanchez-Lengeling of Google Brain for mentorship and guidance; we thank Dr. Ashish Vaswani for technical advice.

\bibliographystyle{plain}
\bibliography{references}

\begin{thebibliography}{10}

\bibitem{noauthor_undated-zw}
bert: {TensorFlow} code and pre-trained models for {BERT}.

\bibitem{noauthor_undated-zv}
Database of perfumery materials \& performance.
\newblock \url{http://www.leffingwell.com/bacispmp.htm}.
\newblock Accessed: 2022-5-6.

\bibitem{noauthor_undated-wj}
The good scents company - flavor, fragrance, food and cosmetics ingredients
  information.
\newblock \url{http://www.thegoodscentscompany.com/}.
\newblock Accessed: 2022-5-6.

\bibitem{noauthor_undated-ir}
January 2022 crawl archive now available -- common crawl.
\newblock
  \url{https://commoncrawl.org/2022/02/january-2022-crawl-archive-now-available/}.
\newblock Accessed: 2022-2-14.

\bibitem{noauthor_undated-mg}
Leather.
\newblock \url{https://www.fragrantica.com/notes/Leather-156.html}.
\newblock Accessed: 2022-2-14.

\bibitem{noauthor_undated-ub}
Perfume and fragrances -- independent guide, reviews, news, chat.
\newblock \url{https://basenotes.com/}.
\newblock Accessed: 2022-5-6.

\bibitem{noauthor_undated-dp}
Perfumes and colognes magazine, perfume reviews and online
  {Community---Fragrantica.com}.
\newblock \url{https://www.fragrantica.com/}.
\newblock Accessed: 2022-5-6.

\bibitem{noauthor_undated-is}
{TPU} research cloud.
\newblock \url{https://sites.research.google/trc/about/}.
\newblock Accessed: 2022-3-27.

\bibitem{Bojanowski2016-og}
Piotr Bojanowski, Edouard Grave, Armand Joulin, and Tomas Mikolov.
\newblock Enriching word vectors with subword information.
\newblock {\em arXiv preprint arXiv:1607. 04606}, 2016.

\bibitem{Brown2020-og}
Tom~B Brown, Benjamin Mann, Nick Ryder, Melanie Subbiah, Jared Kaplan, Prafulla
  Dhariwal, Arvind Neelakantan, Pranav Shyam, Girish Sastry, Amanda Askell,
  Sandhini Agarwal, Ariel Herbert-Voss, Gretchen Krueger, Tom Henighan, Rewon
  Child, Aditya Ramesh, Daniel~M Ziegler, Jeffrey Wu, Clemens Winter,
  Christopher Hesse, Mark Chen, Eric Sigler, Mateusz Litwin, Scott Gray,
  Benjamin Chess, Jack Clark, Christopher Berner, Sam McCandlish, Alec Radford,
  Ilya Sutskever, and Dario Amodei.
\newblock Language models are {Few-Shot} learners.
\newblock May 2020.

\bibitem{Devlin2018-ke}
Jacob Devlin, Ming-Wei Chang, Kenton Lee, and Kristina Toutanova.
\newblock {BERT:} pre-training of deep bidirectional transformers for language
  understanding.
\newblock October 2018.

\bibitem{Dravnieks_A_ASTM_Committee_E-18_on_Sensory_Evaluation_of_Materials_and_Products1985-cx}
{Dravnieks, A. , \& ASTM Committee E-18 on Sensory Evaluation of Materials and
  Products.}
\newblock Atlas of odor character profiles, 1985.

\bibitem{Golovin2017-eh}
Daniel Golovin, Benjamin Solnik, Subhodeep Moitra, Greg Kochanski, John~Elliot
  Karro, and D~Sculley.
\newblock Google vizier: A service for black-box optimization.
\newblock 2017.

\bibitem{Gutierrez2018-hh}
E~Dar{\'\i}o Guti{\'e}rrez, Amit Dhurandhar, Andreas Keller, Pablo Meyer, and
  Guillermo~A Cecchi.
\newblock Predicting natural language descriptions of mono-molecular odorants.
\newblock {\em Nat. Commun.}, 9(1):4979, November 2018.

\bibitem{Honnibal2017-pv}
Matthew Honnibal and Ines Montani.
\newblock {spaCy 2}: Natural language understanding with {B}loom embeddings,
  convolutional neural networks and incremental parsing.
\newblock 2017.

\bibitem{Jiang2019-fd}
Zhengbao Jiang, Frank~F Xu, Jun Araki, and Graham Neubig.
\newblock How can we know what language models know?
\newblock November 2019.

\bibitem{Keller2017-og}
Andreas Keller, Richard~C Gerkin, Yuanfang Guan, Amit Dhurandhar, Gabor Turu,
  Bence Szalai, Joel~D Mainland, Yusuke Ihara, Chung~Wen Yu, Russ Wolfinger,
  Celine Vens, Leander Schietgat, Kurt De~Grave, Raquel Norel, {DREAM Olfaction
  Prediction Consortium}, Gustavo Stolovitzky, Guillermo~A Cecchi, Leslie~B
  Vosshall, and Pablo Meyer.
\newblock Predicting human olfactory perception from chemical features of odor
  molecules.
\newblock {\em Science}, 355(6327):820--826, February 2017.

\bibitem{Keller2016-ci}
Andreas Keller and Leslie~B Vosshall.
\newblock Olfactory perception of chemically diverse molecules.
\newblock {\em BMC Neurosci.}, 17(1):55, August 2016.

\bibitem{Lan2017-nq}
Wuwei Lan, Siyu Qiu, Hua He, and Wei Xu.
\newblock A continuously growing dataset of sentential paraphrases.
\newblock August 2017.

\bibitem{Lester2021-vg}
Brian Lester, Rami Al-Rfou, and Noah Constant.
\newblock The power of scale for {Parameter-Efficient} prompt tuning.
\newblock April 2021.

\bibitem{Liu2021-qu}
Xiao Liu, Yanan Zheng, Zhengxiao Du, Ming Ding, Yujie Qian, Zhilin Yang, and
  Jie Tang.
\newblock {GPT} understands, too.
\newblock March 2021.

\bibitem{Pedregosa2011-mu}
F~Pedregosa, G~Varoquaux, A~Gramfort, V~Michel, B~Thirion, O~Grisel, M~Blondel,
  P~Prettenhofer, R~Weiss, V~Dubourg, J~Vanderplas, A~Passos, D~Cournapeau,
  M~Brucher, M~Perrot, and E~Duchesnay.
\newblock Scikit-learn: Machine learning in {P}ython.
\newblock {\em J. Mach. Learn. Res.}, 12:2825--2830, 2011.

\bibitem{Qin2021-sg}
Guanghui Qin and Jason Eisner.
\newblock Learning how to ask: Querying {LMs} with mixtures of soft prompts.
\newblock April 2021.

\bibitem{Schick2020-ks}
Timo Schick and Hinrich Sch{\"u}tze.
\newblock It's not just size that matters: Small language models are also
  {Few-Shot} learners.
\newblock September 2020.

\bibitem{Shin2020-gy}
Taylor Shin, Yasaman Razeghi, Robert~L Logan, IV, Eric Wallace, and Sameer
  Singh.
\newblock {AutoPrompt}: Eliciting knowledge from language models with
  automatically generated prompts.
\newblock October 2020.

\bibitem{Thoppilan2022-fu}
Romal Thoppilan, Daniel De~Freitas, Jamie Hall, Noam Shazeer, Apoorv
  Kulshreshtha, Heng-Tze Cheng, Alicia Jin, Taylor Bos, Leslie Baker, Yu~Du,
  Yaguang Li, Hongrae Lee, Huaixiu~Steven Zheng, Amin Ghafouri, Marcelo
  Menegali, Yanping Huang, Maxim Krikun, Dmitry Lepikhin, James Qin, Dehao
  Chen, Yuanzhong Xu, Zhifeng Chen, Adam Roberts, Maarten Bosma, Vincent Zhao,
  Yanqi Zhou, Chung-Ching Chang, Igor Krivokon, Will Rusch, Marc Pickett,
  Pranesh Srinivasan, Laichee Man, Kathleen Meier-Hellstern, Meredith~Ringel
  Morris, Tulsee Doshi, Renelito~Delos Santos, Toju Duke, Johnny Soraker, Ben
  Zevenbergen, Vinodkumar Prabhakaran, Mark Diaz, Ben Hutchinson, Kristen
  Olson, Alejandra Molina, Erin Hoffman-John, Josh Lee, Lora Aroyo, Ravi
  Rajakumar, Alena Butryna, Matthew Lamm, Viktoriya Kuzmina, Joe Fenton, Aaron
  Cohen, Rachel Bernstein, Ray Kurzweil, Blaise Aguera-Arcas, Claire Cui,
  Marian Croak, Ed~Chi, and Quoc Le.
\newblock {LaMDA}: Language models for dialog applications.
\newblock January 2022.

\bibitem{Vaswani2017-wh}
Ashish Vaswani, Noam Shazeer, Niki Parmar, Jakob Uszkoreit, Llion Jones,
  Aidan~N Gomez, Lukasz Kaiser, and Illia Polosukhin.
\newblock Attention is all you need.
\newblock June 2017.

\bibitem{Wolf2019-te}
Thomas Wolf, Lysandre Debut, Victor Sanh, Julien Chaumond, Clement Delangue,
  Anthony Moi, Pierric Cistac, Tim Rault, R{\'e}mi Louf, Morgan Funtowicz, Joe
  Davison, Sam Shleifer, Patrick von Platen, Clara Ma, Yacine Jernite, Julien
  Plu, Canwen Xu, Teven Le~Scao, Sylvain Gugger, Mariama Drame, Quentin Lhoest,
  and Alexander~M Rush.
\newblock {HuggingFace's} transformers: State-of-the-art natural language
  processing.
\newblock October 2019.

\bibitem{Zarzo2021-rb}
Manuel Zarzo.
\newblock Multivariate analysis and classification of 146 odor character
  descriptors, 2021.

\end{thebibliography}

\section{Appendix}
\subsection{Attached Extra Files}
improvement\_by\_desc.csv
\subsection{Failed Approaches}
\label{sec:failed}
The two presented techniques, human generated prompts and mined prompts, were only a subset of all techniques we applied to this benchmark. We actually implemented other techniques which resulted in failure. This section enumerates the other techniques and discusses their costs and benefits, especially with regard to their computational cost and ease of implementation, so that future researchers can decide on techniques based on their resources.

Firstly, the most obvious method to improve on BERT off-the-shelf is to fine-tune the weights of the BERT model. The official BERT GitHub\cite{noauthor_undated-zw} contains a script to fine-tune the model, meaning there was little to no implementation work. However when improvements on the BERT MLM task using our dataset did not translate to improvements on the descriptor prediction benchmark, it was unclear as to whether our gathered dataset was low quality, or whether a different training technique was necessary. In addition, because a single trial in the hyperparameter tuning process done through Vizier\cite{Golovin2017-eh} may involve 10-100k training steps, this method is costly, and we needed to spin up a 64-core Cloud TPU\cite{noauthor_undated-is}. One source of our issues may have been that the corpus involved in pre-training BERT is on the scale of billions of words\cite{Devlin2018-ke}, while finding a corpus of even a million words is infeasible for odor right now; a larger corpus may have been necessary to get better results from fine-tuning.

When fine-tuning did not pan out, we looked into prompting as a more flexible and less expensive method. We first attempted human generated prompts, which brought good results, but we looked to previous work to improve further. None of these techniques require a training dataset, and we were able to use the benchmark score as the direct evaluation metric.

We attempted to tune a prompt of fixed size consisting of tokens sampled from BERT’s vocabulary, originally presented by "AutoPrompt: Eliciting knowledge from language models..."\cite{Shin2020-gy} and then optimized in "Learning How to Ask: Querying {LMs}..."\cite{Qin2021-sg}. Essentially, a number of tokens are selected as the initial prompt, and then the prompt tokens are permuted and replaced over the course of training. We attempted to use the techniques described in both papers, and while the only hyperparameter was prompt length, their source codes did require a decent amount of modification to work with our project. In the end, we did not get usable results from either approach. 

We also tried tuning a vector prompt of fixed size, based on work presented in "The Power of Scale for Parameter-Efficient Prompt Tuning"\cite{Lester2021-vg}. This technique leverages backpropagation to train a prompt vector fed directly into the model’s transformer layers. This involved more hyperparameters, requiring tuning through Vizier. As above, using this technique did not improve meaningfully on BERT off-the-shelf.

\begin{table}[h]
\small
\begin{tabular}{ |p{2.5cm}|p{2.5cm}|p{2.5cm}|p{2.5cm}|}
	\hline
	Technique                & \small \makecell[tl]{ Implementation \\ Cost (weeks)} & \small \makecell[tl]{ Training Cost \\ (predictions)} & \small Corpus required? \\
	\hline
	\makecell[lt]{BERT \\ (off-the-shelf)}       & None                        & None                        & No               \\
	\hline
	\makecell[lt]{BERT \\ (fine-tuning)}          & Less than one week          & 100M+                       & Yes (Large)      \\
	\hline
	\makecell[tl]{ Human prompt}   & Less than one week          & ~15                         & No               \\
	\hline
	\makecell[tl]{ Fixed-sized \\ token prompt} & Two weeks                   & 10k                         & No               \\
	\hline
	Numerical prompt         & Three weeks                 & 500k                        & No               \\
	\hline
	Mined prompt             & Two weeks                   & 250k                        & Yes              \\
	\hline
\end{tabular}
\caption{\label{tab:costs} \textbf{Cost analysis} of the fine-tuning and prompting methods we explored. }
\end{table}

Future researchers would benefit from following Occam’s razor and leveraging the simplest approach that meets their needs first before exploring others. 

\end{document}